\documentclass{IEEEtran} 
\usepackage{hyperref}
\usepackage{algorithmic}
\usepackage{algorithm}

\usepackage{color}
 
\usepackage{cite} 
\usepackage{amsmath,amsfonts}
\usepackage{enumerate} 
\usepackage{booktabs}

\newtheorem{lemma}{Lemma} 
\newtheorem{theorem}{Theorem} 
\newtheorem{definition}{Definition} 
\newtheorem{example}{Example}   
\newtheorem{corollary}{Corollary}
\newcommand{\opt}{\mathrm{opt}} 

\begin{document}

\title{A Theoretical Framework of Approximation Error Analysis of Evolutionary Algorithms}

\author{Jun He, Yu Chen  and Yuren Zhou
\thanks{This work was supported by EPSRC under Grant No. EP/I009809/1}
\thanks{Jun He is with the School of Science and Technology, Nottingham Trent University, Clifton Campus, Nottingham NG11 8NS, UK}
\thanks{Yu Chen is with the School of Science, Wuhan University of Technology, Wuhan, 430070, China.}
\thanks{Yuren.. Zhou is with the School of Data and Computer Science,
Sun Yat-sen University, Guangzhou 510006, China.}
}

\maketitle

\begin{abstract} 
In  the empirical study of evolutionary algorithms, the solution quality is evaluated by either the fitness value or approximation error. The latter measures the fitness difference between an approximation solution and the optimal solution. Since  the  approximation error analysis is more convenient than the direct estimation of the fitness value, this paper focuses on  approximation error analysis. However, it is straightforward to extend all related results from the approximation error  to the fitness value.  Although the evaluation of solution quality plays an essential role in practice, few rigorous analyses have been conducted on this topic. This paper aims at establishing a novel theoretical framework of approximation error analysis of evolutionary algorithms for discrete optimization. This  framework is divided into two parts. The first part is about exact expressions of the approximation error. Two methods, Jordan form and Schur's triangularization, are presented to obtain an exact expression. The second part is about upper bounds on approximation error. Two methods, convergence rate and  auxiliary matrix iteration,  are proposed to estimate the upper bound. The applicability of this framework is demonstrated through several examples.
\end{abstract}


\section{Introduction}
\label{sec:introduction}
In  the empirical study of evolutionary algorithms (EAs), the quality of a solution is evaluated by either the fitness value or approximation error. The latter measures the fitness difference between   an approximation solution and the optimal solution. The absolute error of a solution $X$ is defined by  $|f(X)-f_{\opt}|$ where $f_{\opt}$ is the fitness of the optimal solution and $f(X)$  the fitness of  $X$~\cite{he2016analytic,he2017initial}.  The approximation error has been widely used in the empirical study of EAs in either a standard form $|f(X)-f_{\opt}|$ or its logarithmic scale $\log |f(X)-f_{\opt}|$~\cite{auger2005performance,qu2013distance, sarker2014differential,wang2014differential, guo2015enhancing, tang2015differential, maesani2016memetic,sinha2018using}. Starting from the absolute error $|f(X)-f_{\opt}|$, it is trivial to derive the fitness value $f(X)= f_{\opt} \pm |f(X)-f_{\opt}|$ where $+$ for a minimization problem and $-$ for a maximization problem. Therefore, this paper focuses on analyzing the approximation error of EAs. It is straightforward to extend related results from the approximation error  to the fitness value of EAs.

Although the fitness value or approximation error has been widely  adopted to evaluate the performance of EAs in computational experiments, they are seldom studied in a rigorous way. This is in shark contrast to the computational time of EAs. The latter is today's mainstream in the theory of EAs~\cite{oliveto2007time} but in the practice, computational time is seldom  applied to evaluating the performance of EAs. In order to bridge this gap between practice and theory, it is necessary to make a rigorous error analysis of EAs. 

Because EAs are random iterative algorithms,  the  expected value, $e^{[t]}=\mathbb{E}[ |f(X^{[t]})-f_{\opt}|]$ of the $t$th generation solution $X^{[t]}$,  is a function of $t$. The main research questions are two questions: (1) what is an exact expression of $e^{[t]}$? (2) if an exact expression is unavailable, what is a bound on $e^{[t]}$?   He~\cite{he2016analytic} made one of the first attempts to answer these questions. He gave an analytic expression of the approximation error for a class of (1+1) strictly elitist EAs. 

This paper aims at establishing a theoretical framework of studying  approximation error of EAs for discrete optimization. 
In the framework, EAs are modelled by homogeneous Markov chains. The analysis is divided into two parts. The first part is about exact expressions of the approximation error. Two methods, Jordan form and Schur’s triangularization, are given to study the exact expression of $e^{[t]}$. The second part is about upper bounds on the approximation error.  
Two methods, convergence rate and auxiliary matrix iteration,   are introduced to the estimatation of the upper bound on $e^{[t]}$.

The paper is arranged as follows: Section~\ref{secLinks} reviews  links to related work.  Section~\ref{secPreliminary} presents preliminary definitions,  notation and Markov modelling of EAs. Section~\ref{secExact} demonstrates the exact expression of   approximation error. Section~\ref{secUpper} estimates the upper bound on   approximation error.  Section~\ref{secConclusions} summarizes the paper.

\section{Related Work}
\label{secLinks} 
In  practice,  approximation error has been widely used to evaluate the quality of solutions found by EAs~\cite{auger2005performance,qu2013distance, sarker2014differential,wang2014differential, guo2015enhancing, tang2015differential, maesani2016memetic,sinha2018using}.  When evaluating  the performance of EAs, we   list  solution  error  in a table or display error trend in a figure. Then we claim that the algorithm with the smallest $e^{[t]}$ value is the best one at the $t$th generation.  Approximation error is called in different names,  such as,   objective function error~\cite{auger2005performance}, difference from a computed solution to a known global optimum~\cite{qu2013distance}, distance from the optimum~\cite{maesani2016memetic,sinha2018using}, fitness error~\cite{tang2015differential} or solution error~\cite{wang2014differential,guo2015enhancing}.

So far, the theoretical study of approximation error  is rare in evolutionary computation. Rudolph~\cite{rudolph1997convergence} proved that under the condition $e^{[t]}/e^{[t-1]} \le \lambda <1$, the sequence $\{e^{[t]}; t=0,1, \cdots\}$ converges in mean geometrically fast to $0$, that is, $\lambda^t e^{[t]}=o(1)$.   

Recently He~\cite{he2016analytic} made one of the first attempts to  obtain an analytic expression of the approximation error for a class of elitist EAs. He proved if
the transition matrix associated with an EA is an upper triangular matrix with unique diagonal entries, then for any $t \ge 1$, the relative  error $e^{[t]}$ is expressed by
$ e^{[t]} =\sum^{L}_{k=1}  c_k \lambda_k^{t-1},
$ where $\lambda_k$  are eigenvalues of the transition matrix (except the largest eigenvalue $1$) and $c_k$ are coefficients. 

He and Lin~\cite{he2016average} studied the geometric average convergence rate of the error sequence $\{e_t; t=0,1, \cdots\}$, defined by 
\begin{align}
R^{[t]} = 1- \left( \frac{e^{[t]}}{e^{[0]}} \right)^{1/t}.
\end{align}   
Starting from $R^{[t]}$, it is straightforward to draw an exact expression of the approximation error: $e^{[t]} = (1-R^{[t]})^t e^{[0]}$. They estimated the lower bound on $R^{[t]}$ and proved if the initial population is sampled at random, $R^{[t]}$ converges to an eigenvalue of the transition matrix associated with an EA.

A close work is fixed budget analysis proposed by Jansen and Zarges~\cite{jansen2012fixed,jansen2014performance}. They aim to bound the fitness value $f(X^{[t]})$ within a fixed time budget. The obtained bounds usually hold within some fixed $t$. For example, the lower and upper bounds given in~\cite[Theorem 9]{jansen2014performance}  are expressed in the form $c_1 t-c_0\le \mathbb{E}[f(X^{[t]})] \le c'_1 t-c'_0$ for some fixed  $t$. However, when $t \to +\infty$, these lower and upper bounds go towards $+\infty$; thus they become invalid  bounds on $f^{[t]}$ for large $t$. This observation reveals an essential difference between fixed budget analysis  and approximation error analysis. In fixed budget analysis,  a bound is an approximation of $f(X^{[t]})$ for some small $t$ but might be invalid for large $t$. The expression of bounds could be a linear   or  exponential function of $t$. But approximation error analysis proves that $e^{[t]}$   always can be upper-bounded by exponential functions of $t$. The bound is valid for all $t$. In this sense, approximation error analysis may be called \emph{any budget analysis}.

\section{Preliminary}
\label{secPreliminary} 
\subsection{Definitions and Notation} 
\label{secDefinition}
We consider a maximization problem: 
\begin{align}
\max f(x), \mbox{ subject to } x \in \mathcal{S},
\end{align}
where $f(x)$ is a fitness function such that $\mid f(x) \mid<+\infty$ and its definition domain $\mathcal{S}$ is a finite state set. Let $f_{\opt}$ denote the maximal value of $f$ and $S_{\opt}=\{x \mid f(x)=f_{\opt}\}$ the optimal solution set.

In evolutionary computation, an individual is a solution $x \in \mathcal{S}$. A population  is a collection of individuals. Let $\mathcal{P}$ denote the  population set.  The fitness of a population $X$ is   $f(X) =\max \{f(x) \mid x \in X \}.$ 

A general EA for solving the above optimization problem is described in Algorithm~\ref{alg1}.  The EA is stopped once an optimal solution is found. This stopping criterion is taken for the sake of theoretical analysis. An EA is called \emph{elitist (or strictly elitist)} if $f(X^{[t]}) \ge (>) f(X^{(t-1)})$ for any $t$. Any non-elitist EA can be modified into an
equivalent elitist EA through adding an archive individual
which preserves the best found solution but does not get
involved in evolution.  

\begin{algorithm}[ht]
\caption{A general   EA}
\label{alg1}
\begin{algorithmic}[1]
\STATE counter $t \leftarrow 0$;
\STATE population $X^{(0)} \leftarrow$ initialize a population of solutions;
\WHILE{$f(X^{[t]}) < f_{\opt}$}
\STATE population $X^{(t+1)}\leftarrow$ apply genetic operators (mutation, crossover, selection or other operators) on $X_{t}$;
\STATE counter $t\leftarrow t+1$;
\ENDWHILE
\end{algorithmic}
\end{algorithm}

\begin{definition} Given an initial state $X^{[0]}$, the \emph{fitness of $X^{[t]}$} is denoted by $f(X^{[t]}\mid X^{[0]})$ (or $f(X^{[t]} )$ in short thereafter) and its expected value is 
$
f^{[t]}  =\mathbb{E}[f(X^{[t]} )].
$
The \emph{absolute  error} of $X^{[t]}$ is  $e(X^{[t]}) = \mid  f(X^{[t]}  )-f_{\opt}\mid$ and its expected value is
$e^{[t]} =\mathbb{E}[e(X^{[t]} )].$
An EA is called to \emph{converge in mean} if $\lim_{t \to +\infty} e^{[t]}=0$ for any  $X^{[0]}$.
\end{definition}

\subsection{Transition Matrix} 
\label{secMarkov}
The approximation error analysis of EAs is built upon the Markov chain modelling of EAs which can be found in existing references such as~\cite{suzuki1995markov,schmitt2001importance,he2003towards}.
A similar Markov chain framework has been used to analyze the computational time of EAs in~\cite{he2003towards}. This paper focuses on a different topic, the approximation error of EAs. 

For the sake of notation, population states are indexed by $\{ 0, 1, \cdots, L\}$. The index $0$ represents the set  of optimal populations. Other indexes $1, \cdots, L$ represent  non-optimal  populations.  Populations are sorted according to their fitness value from high to low: 
\begin{align*}
f_{\max} =f_0  > f_1 \ge \cdots \ge f_L = f_{\min},
\end{align*} 
where $f_i$ stands for $f(i)$ in short. The decomposition of states is  not required to satisfy $f_0  > f_1 > \cdots > f_L$.  Examples~\ref{exampleNeedle1} and \ref{exampleNeedle2} in  next section will show this point.  

The sequence $\{X^{[t]}; t=0,1,\cdots \}$ is a  Markov chain because $X^{[t]}$ is determined by $X^{[t-1]}$ in a probability way. Furthermore we assume that  the transition probability from any $i$ to   $j$ doesn't change over $t$. So the chain is homogeneous. The transition probability from $j$ to $i$ is denoted by 
\begin{align}
p_{i,j}=\Pr(X^{[t]}= i \mid  X^{[t-1]} = j),  &&  i,j=0, \cdots, L.
\end{align}

Let  $\mathbf{v}$ stand for a column vector and $\mathbf{v}^T$ for the row column with the transpose operation $T$. The transition  matrix  is a $(L+1) \times (L+1)$ matrix: 
\begin{align} 
\mathbf{P}=
\begin{pmatrix}
1&\mathbf{r}^T\\
\mathbf{0}&\mathbf{R}
\end{pmatrix}
\end{align}
where $p_{0,0}=1$ is due to the stopping criterion. The vector $\mathbf{r}$ denotes transition probabilities from non-optimal states  to optimal ones. The zero-valued vector $\mathbf{0}$ means that transition probabilities from  optimal states to non-optimal ones are 0. The matrix $\mathbf{R}$ represents transition probabilities within non-optimal states, given by 
\begin{align}
\label{matrixR}
\mathbf{R}=&\begin{pmatrix}  
  p_{1,1}    & p_{1,2}& p_{1,3}& \cdots &   p_{1,L-1}  &p_{1,L}  \\ 
    p_{2,1} & p_{2,2} & p_{2,3}& \cdots & p_{2,L-1} &p_{2,L}\\
 p_{3,1}  &  p_{3,2} & p_{3,3} &  \cdots & p_{3,L-1} &p_{3,L}\\
 \vdots & \vdots &  \vdots  & \vdots & \vdots    \\
  p_{L,1} &  p_{L,2} &  p_{L,3} &\cdots & p_{L,L-1} & p_{L,L}     \\
 \end{pmatrix}. 
\end{align}

The error sequence $\{e^{[t]};t=0,1,\cdots \}$ can be written by a matrix iteration. Then we get the first exact expression of $e^{[t]}$.

\begin{theorem}
\label{theorem1} Let $e(i)$ (or $e_i$ in short) denote the  approximation error of  $i$:  $e(i)=|f(i)-f_{\opt}|$ and $\mathbf{e}^T =
 (e_1 , \cdots, e_L). $ $\mathbf{p}^{[0]}$ denotes the probability distribution of $X^{[0]}$ over non-optimal states $\{1, \cdots, L\}$.  
Then
\begin{align}
\label{equApproximationError} 
 e^{[t]}  = \mathbf{e}^T \mathbf{R}^t  \mathbf{p}^{[0]}.
\end{align}   
\end{theorem}

\begin{IEEEproof}
Let   $p^{[t]}(i)$ (or $p^{[t]}_i$ in short) denote the probability $\Pr(X^{[t]}=i)$. Because $e_0=0$, we have 
\begin{align}
e^{[t]}= \textstyle \sum^L_{i=0} p^{[t]}_i e_i = \sum^L_{i=1} p^{[t]}_i e_i.
\end{align} 

According to the Markov chain property, for any $i \neq 0$,  
\begin{align}
\label{equDistribution}
p^{[t]}_i =\textstyle   \sum^L_{j=0}p_{i,j} p^{[t-1]}_j =\sum^L_{j=1}p_{i,j} p^{[t-1]}_j.
\end{align}
Let $
\mathbf{p}^{[t]}: = ( p^{[t]}_1,  \cdots, p^{[t]}_L)^T$. (\ref{equDistribution}) is rewritten as
\begin{align}
&\mathbf{p}^{[t]}= \mathbf{R} \mathbf{p}^{[t-1]}    = \mathbf{R}^t \mathbf{p}^{[0]} . 
\label{equMatrixIteration}  
\end{align}
 Then we get $ e^{[t]}  = \mathbf{e}^T \mathbf{R}^t  \mathbf{p}^{[0]}$.
\end{IEEEproof}

The above theorem shows   that $e^{[t]}$ is determined by   $\mathbf{p}^{[0]}$,   matrix power $\mathbf{R}^t$ and    $\mathbf{e}^T$. Only $\mathbf{R}^t$ changes over   $t$, thus it plays the most important role in expressing  $e^{[t]}$. (\ref{equApproximationError}) also reveals it is sufficient to use partial transition matrix $\mathbf{R}$, rather than the whole transition matrix $\mathbf{P}$ for expressing $e^{[t]}$.

\subsection{Matrix Analysis}   
Matrix analysis is the main mathematical tool used in  the error analysis of EAs. Several essential definitions and lemmas are listed here. Their details can be found in the textbook~\cite{meyer2000matrix}.

\begin{definition}
For an $L \times L$ matrix $\mathbf{A}$, scalars $\lambda$ and $L\times 1$ vectors $\mathbf{v} \neq 0$ satisfying $\mathbf{A}\mathbf{v} = \lambda \mathbf{v}$ are called eigenvalues and eigenvectors of $\mathbf{A}$ respectively. 
A complete set of eigenvectors for $\mathbf{A}$ is any set of $L$ linearly
independent eigenvectors for $\mathbf{A}$. Let $\lambda_{\max}=\max \{|\lambda_1|, \cdots, |\lambda_L|\}$, which is called the spectral radius of  matrix $\mathbf{A}$.
\end{definition}

\begin{definition}
A matrix $\mathbf{A}$ is called \emph{diagonalizable} if there exists a matrix $\mathbf{Q}$ such that $\mathbf{A}=\mathbf{Q}^{-1} \mathbf{D}\mathbf{Q}$ where $\mathbf{D}$ is diagonal matrix with diagonal entries $\lambda_i$ and $\lambda_i$ is an eigenvalue of $\mathbf{A}$. 
\end{definition}

\begin{lemma}
\label{lemmaDiagonal}
A square matrix $\mathbf{A}$ is diagonalizable if and only if $\mathbf{A}$ possesses a complete set of eigenvectors.
\end{lemma}

\begin{definition}
A unitary matrix is defined to be a $L \times L$ complex matrix $\mathbf{U}$ whose columns (or rows) constitute an orthonormal basis for $\mathbb{C}^L$. 
\end{definition}

\begin{lemma}
\label{lemma3}
$\mathbf{A}$ is real symmetric if and only if $\mathbf{A}$ is orthogonally similar to a
real-diagonal matrix $\mathbf{D}$, that is, $\mathbf{Q}^T \mathbf{A}\mathbf{Q} = \mathbf{D}$ for some orthogonal $\mathbf{Q}$.
\end{lemma}

\begin{lemma}[Schur’s Triangularization]
\label{lemmaSchur}
Every square matrix is unitarily similar to an upper-triangular matrix.
That is, for each $\mathbf{A}$, there exists a unitary matrix $\mathbf{U}$ (not unique)
and an upper-triangular matrix $\mathbf{T}$ (not unique) such that $\mathbf{U}^* \mathbf{A} \mathbf{U}=\mathbf{T}$,
and the diagonal entries of $\mathbf{T}$ are the eigenvalues of $\mathbf{A}$.
\end{lemma}

\begin{lemma}[Jordan Form]
\label{lemmaJordan} For every $L \times L$ matrix $\mathbf{A}$ with distinct eigenvalues $\{\lambda_1,  \cdots, \lambda_k\}$, there is  a non-singular matrix $\mathbf{Q}$ such that 
\begin{align} 
\mathbf{A} = \mathbf{Q}^{-1} \mathbf{J} \mathbf{Q}= 
&\begin{pmatrix}  
\mathbf{J}_1 &              &        &  \\ 
             & \mathbf{J}_2 &        & \\ 
             &              &\ddots  &   \\
             &              &        & \mathbf{J}_k      \\
 \end{pmatrix}.
\end{align}
Each Jordan block $\mathbf{J}_i$ is a square matrix of the form
\begin{align} 
\mathbf{J}_i =  
&\begin{pmatrix}  
\lambda_i & 1         &      &           &                \\ 
          & \lambda_i & 1    &           &                 \\ 
          &           &\ddots& \ddots    &               \\
          &           &      & \lambda_i &1    \\ 
          &           &      &           & \lambda_i  \\
 \end{pmatrix} 
\end{align}
where $\lambda_i$ is  an eigenvalue of $\mathbf{A}$.  
Each Jordan block $\mathbf{J}_i$ is a $L_i \times L_i$ square matrix and $L=L_1+\cdots + L_k$.  
\end{lemma}

\section{Exact Expressions of Approximation Errors}
\label{secExact} 
In the error analysis of EAs, the perfect goal is to seek an exact expression of $e^{[t]}$. This section discusses this topic.

\subsection{Jordan Form Method}
\label{secJordan}
Let's start  from  a simple case that transition matrix $\mathbf{P}$ is diagonalizable. We can obtain an exact expression of $e^{[t]}$ as follows.    

\begin{theorem}
\label{theorem2}
If matrix $\mathbf{R}$ is diagonalizable such that $\mathbf{R}=\mathbf{Q}^{-1} \mathbf{D}\mathbf{Q}$ where matrix $\mathbf{D}$ is diagonal matrix,  then   
\begin{align}
e^{[t]} =\textstyle \sum^L_{i=1} c_i \lambda^t_i,
\end{align}
where $\lambda_i$ denote its $i$th diagonal entry of $\mathbf{D}$, $c_i= \sum^L_{i=1} a_i b_i$, vectors  $\mathbf{a}^T=\mathbf{e}^T\mathbf{Q}^{-1}$ and $\mathbf{b}=\mathbf{Q} \mathbf{p}^{[0]}$. 
\end{theorem}

\begin{IEEEproof}
From Theorem~\ref{theorem1}, we know $ e^{[t]}   
 = \mathbf{e}^T \mathbf{R}^t  \mathbf{p}^{[0]}.$ Since $\mathbf{R}=\mathbf{Q}^{-1} \mathbf{D}\mathbf{Q}$, we get $ e^{[t]}  =\mathbf{a}^T   \mathbf{D}^t    \mathbf{b}.$ Since $\mathbf{D}^t$ is a diagonal matrix whose diagonal entries are $\lambda^t_i$, we come to the conclusion.
\end{IEEEproof}

This theorem claims that $e^{[t]}$ is a linear combination of exponential functions $\lambda^t_i$ provided that matrix $\mathbf{R}$ is diagonalizable. Thus, the error analysis of EAs is how to calculate or estimated eigenvalues $\lambda_i$ and coefficients $c_i$.

\begin{example} [EA-BWSE on Needle-in-Haystack]
\label{exampleNeedle1}
Consider the problem of maximizing the Needle-in-Haystack function,  
\begin{align*}  
&\max f(x)= \left\{
\begin{array}{cc}
    1, & \mbox{if }|x|=0, \\
    0, & \mbox{otherwise,}
\end{array} 
\right.  
\end{align*}  
where   $x=(x_1, \cdots, x_n)\in \{0,1\}^n$  and $|x| = x_1+ \cdots +x_n$.  

EA-BWSE, a (1+1) EA with bitwise mutation and strictly elitist selection (Algorithm~\ref{EA-BWSE}),   is used for solving the above maximization problem.

\begin{algorithm}[ht]
\caption{EA-BWSE}
\label{EA-BWSE}
\begin{algorithmic} 
\STATE \textbf{Bitwise Mutation}: flip each bit of $x$ at random and generate $y$;
\STATE \textbf{Elitist Selection}: if $f(y)>f(x)$, then $y$ replace $x$.
\end{algorithmic}
\end{algorithm}

Let index $i$ denote the state of $x$ such that $|x|=i$ where $i=0,1, \cdots, n$. Then transition probabilities satisfy
\begin{align}
\begin{array}{lll}
    &p_{0,0}=1, \\
    & p_{0,i}=\left(\frac{1}{n}\right)^{i} \left(1-\frac{1}{n}\right)^{n-i},\\
    &p_{i,i}=1-\left(\frac{1}{n}\right)^{i} \left(1-\frac{1}{n}\right)^{n-i}.
\end{array}
\end{align}
Transition matrix $\mathbf{P}$ is diagonal. Let $\mathbf{p}^{[0]}$ denote the initial distribution of $X^{[0]}$.  According to Theorem~\ref{theorem2}, the approximation error  
\begin{align}
    e^{[t]}= \textstyle  \sum^n_{i=1} \left[ 1-\left(\frac{1}{n}\right)^{i} \left(1-\frac{1}{n}\right)^{n-i}\right]^t p^{[0]}_i.
\end{align} 
\end{example}

\begin{example} [EA-BWNE on Needle-in-Haystack]
\label{exampleNeedle2}
Consider the problem of maximizing  the Needle-in-Haystack function using EA-BWNE, the (1+1) EA with bitwise mutation and non-strictly elitist selection (Algorithm~\ref{EA-BWNE}).   

\begin{algorithm}[ht]
\caption{EA-BWNE}
\label{EA-BWNE}
\begin{algorithmic} 
\STATE \textbf{Bitwise Mutation}: flip each bit of $x$ at random and generate $y$;
\STATE \textbf{Elitist Selection}: if $f(y)\ge f(x)$, then $y$ replace $x$.
\end{algorithmic}
\end{algorithm} 

Let index $i$ denote the state of $x$ such that the conversion of $x$ from binary to decimal is  $i$ where $i=0,1, \cdots, 2^n$. Then transition probabilities satisfy
\begin{align} 
    &p_{0,0}=1, && p_{i,j}=p_{j,i}, && \forall{i,j \neq 0}
\end{align}
Since transition matrix $\mathbf{R}$ is symmetric, it is diagonalizable. 
According to Theorem~\ref{theorem2}, 
$
e^{[t]} =\textstyle \sum^L_{i=1} c_i \lambda^t_i.
$
Theorem~\ref{theorem2} reveals that $e^{[t]}$ is a linear combination of exponential functions $\lambda^t_i$. However, it is still difficult to calculate  eigenvalues $\lambda_i$ and coefficients $c_i$ due to the difficulty in obtaining $\mathbf{Q}$ and $\mathbf{Q}^{-1}$.
\end{example}

No matter whether matrix $\mathbf{R}$ is diagonalizable or not, it can be represented by a Jordan form. 
Previously the method of Jordan form was used to bound  the probability distribution of solutions co verging towards a stationary distribution~\cite{suzuki1995markov,schmitt2001importance}, that is, $\parallel \mathbf{p}^{[t]}-  \mathbf{p}^{\infty} \parallel_1$ where $\mathbf{p}^{[t]}=[p^{[t]}_i]_{i=0, \cdots,L}$ and $\mathbf{p}^{\infty}$ is the limit of $\mathbf{p}^{[t]}$.
Suzuki~\cite{suzuki1995markov}   derived a lower
bound on $\parallel \mathbf{p}^{[t]}-  \mathbf{p}^{\infty} \parallel_1$ for   simple genetic algorithms through analysing eigenvalues   of the  transition matrix.   Schmitt and Rothlauf~\cite{schmitt2001importance} found that the
convergence rate of $\parallel \mathbf{p}^{[t]}-  \mathbf{p}^{\infty} \parallel_1 \to 0$  is determined by the spectral radius of matrix $\mathbf{R}$ .

In the current paper, we aim to derive an exact expression of $\mathbf{e}^T \mathbf{R}^t  \mathbf{p}^{[0]}$ using the Jordan form method.

\begin{lemma} 
\label{lemma5}
Let  $ \mathbf{R}  = \mathbf{Q}^{-1} \mathbf{J} \mathbf{Q}$ be the Jordan form of  $\mathbf{R}$. Then 
\begin{align} 
\label{equExactError1}
 e^{[t]} 
 =      \mathbf{e}^T \mathbf{Q}^{-1} \mathbf{J}^t \mathbf{Q} \mathbf{p}^{[0]} .
\end{align}  
\end{lemma}

\begin{IEEEproof}
From Jordan form: $\mathbf{R}  = \mathbf{Q}^{-1}\mathbf{J} \mathbf{Q}$, we get $\mathbf{R}^t =\mathbf{Q}^{-1} \mathbf{J}^t \mathbf{Q}$. Inserting this expression into  (\ref{equApproximationError}), we get the desired conclusion.
\end{IEEEproof}

From (\ref{equExactError1}), we see that in order to obtain an exact expression of  $e^{[t]}$, we need to represent $\mathbf{J}^t$. This is given in the following theorem.

\begin{theorem}
\label{theorem4}
For any matrix $\mathbf{R}$, the approximation error
\begin{align}
\label{equExactError2}
 \textstyle e^{[t]}  
=   \sum^k_{i=1} \sum^{L_i}_{m=1}   c_{i_m} \binom{t}{L_i-m+1} \lambda^{t-m+1}_{i},
\end{align}
where the coefficient  
\begin{align}
\label{equExactCoefficient} 
c_{i_m}=& \textstyle \sum^{L_i-m+1}_{j=1} a_{i_j} b_{i_{j-m+1}},& m=1, \cdots, L_i, 
\end{align} and  $a_{i_j}$ and $b_{i_j}$ are given by (\ref{equCoeff0}). Let the binomial coefficient $\binom{i}{j}=0$ if $i<j$. 
\end{theorem}

\begin{IEEEproof}
We assume that matrix  $\mathbf{J}^t$ consists of $k$ Jordan blocks
\begin{align}   
&\begin{pmatrix}  
  \mathbf{J}_1^t    &   &   &  \\ 
    & \mathbf{J}_2^t   &   & \\ 
   &   & \ddots  &    \\
 &    &    & \mathbf{J}_k^t      \\
 \end{pmatrix}.
\end{align}

Let $\mathbf{a}^T=  \mathbf{e}^T \mathbf{Q}^{-1} $ and write it into $\mathbf{a}^T=  (\mathbf{a}^T_1, \mathbf{a}^T_2, \cdots, \mathbf{a}^T_k) $. Let $\mathbf{b} =\mathbf{Q} \mathbf{p}^{[0]} $ and write it into $\mathbf{b}= (\mathbf{b}_1, \mathbf{b}_2, \cdots, \mathbf{b}_k)$. Then  (\ref{equExactError1}) can be rewritten as 
\begin{align} 
\label{equJordanBlock}
e^{[t]}    
 = \textstyle \sum^k_{i=1}
   \mathbf{a}^T_i \mathbf{J}^t_i \mathbf{b}_i   
\end{align}

Denote vectors
\begin{align}
\label{equCoeff0}
\mathbf{a}^T_i=(a_{i_1}, \cdots, a_{i_{L_i}}), &&
\mathbf{b}_i=(b_{i_1}, \cdots, b_{i_{L_i}})^T.
\end{align}

Consider the component $\mathbf{a}^T_i \mathbf{J}^t_i \mathbf{b}_i$ in   (\ref{equJordanBlock}). 
Each Jordan block power $\mathbf{J}_i^t$ equals to~\cite[pp. 618]{meyer2000matrix}
\begin{align*}  
\mathbf{J}_i^t=&\begin{pmatrix}  
\lambda_i^t & \binom{t}{1}\lambda_i^{t-1} &\binom{t}{2} \lambda_i^{t-2} &\cdots & \binom{t}{L_i} \lambda_i^{t-L_i}  \\ 
            & \lambda_i^t                 &\binom{t}{1} \lambda_i^{t-1} &\cdots & \binom{t}{L_i-1} \lambda_i^{t-L_i+1}  \\ 
            &                             & \ddots                      &  \ddots &\vdots  \\
                        &                             &                      &  \ddots & \binom{t}{1}\lambda_i^{t-1}  \\
            &                             &                             &         & \lambda_i^{t}  \\ 
 \end{pmatrix}.
\end{align*}
Inserting it into $ \mathbf{a}^T_i \mathbf{J}^t_i \mathbf{b}_i$, we get that  $ \mathbf{a}^T_i \mathbf{J}^t_i \mathbf{b}_i$ equals to
\begin{align}
 \sum^{L_i}_{m=1} \sum^{L_i-m+1}_{j=1} a_{i_j} b_{i_j-m+1} \binom{t}{L_i-m+1}  \lambda^{t-m+1}_{i}.
  \end{align}
Then we have
\begin{align}
\mathbf{a}^T_i \mathbf{J}^t_i \mathbf{b}_i  
=  \sum^{L_i}_{m=1}   c_{i_m} \binom{t}{L_i-m+1} \lambda^{t-m+1}_{i}, 
  \end{align}
where coefficients $c_{i_m}$ is given by (\ref{equExactCoefficient}).
  
The approximation error is the summation of all $i$ from $1$ to $k$, which equals to
\begin{align} 
e^{[t]}  
=   \sum^k_{i=1} \sum^{L_i}_{m=1}   c_{i_m} \binom{t}{L_i-m+1} \lambda^{t-m+1}_{i}.
\end{align} 
The above is the desired result.
\end{IEEEproof}

Theorem~\ref{theorem4} reveals  the exact expression of $e^{[t]}$   consisting of three parts:
\begin{enumerate}
\item \textbf{Exponential terms $\lambda^{t-m+1}_{i}$.} Each term is an exponential function of $t$ where each $\lambda_i$ is an eigenvalue of $\mathbf{R}$.

\item \textbf{Constant Coefficients $c_{i_m}$.} They are independent of $t$. (\ref{equExactCoefficient}) shows that they are determined by vectors $\mathbf{a}^T=\mathbf{e}^T \mathbf{Q}^{-1} $, $\mathbf{b}=\mathbf{Q} \mathbf{p}^{[0]}$  and  the size of Jordan block $L_i$. 

\item \textbf{Binomial coefficients $\binom{t}{L_i-m+1}$.} 
Since $m \le L_i\le L$, each coefficient is a polynomial function of $t$ and its order is up to $t^L$. Binomial coefficients are only related to the size of Jordan block $L_i$.
\end{enumerate}

Because of the difficulty of obtaining Jordan form  of transition matrices, it is hard to generate an exact expression of $e^{[t]}$ in practice.  

As a direct consequence  of Theorem~\ref{theorem4}, we get the sufficient and necessary condition of convergence of EAs.
\begin{corollary}
$\lim_{t \to \infty} e^{[t]}=0$ if and only if $\lambda_{\max}<1$.
\end{corollary}

\subsection{Shur's Decomposition Method}
Alternately matrix power $\mathbf{R}^t$ can be represented using Schur's triangularisation. Then we obtain another exact expression of  $e^{[t]}$. 
Let's start from a simple case that 
matrix $\mathbf{R}$ is upper triangular with distinct eigenvalues $\lambda_1, \cdots, \lambda_L$. The analysis is based on power factors of a matrix~\cite{shur2011simple}.

\begin{definition} 
\label{defMain} 
For an upper triangular matrix  $\mathbf{A}$,  its power factors, $[p_{i,j,k}]$ (where $i,j,k=1, \cdots, L$), are  defined as follows:
\begin{align} 
\label{equPowerFactors}
p_{i,j,k}=\left\{\begin{array}{lll}
 a_{j,j},  &\textrm{if }i=j=k, 
\\
 0,                &\textrm{if }k<i \mbox{ or } k >j, 
\\
 \frac{\sum^{j-1}_{l=k} p_{i,l,k} a_{l,j}}{a_{k,k}-a_{j,j}}, &\textrm{if }i \le k< j, 
\\
a_{i,j}-\sum^{j-1}_{l=i} p_{i,j,l}, &\textrm{if }i<j \textrm{ and } j=k. 
\end{array}
\right.
\end{align}
\end{definition}
 
Using power factors of $\mathbf{R}$, we can obtain an explicit expression of the approximation error $e^{[t]}$ as shown in the theorem below.

\begin{theorem}
\label{theorem5}
If  matrix $\mathbf{R}$ is upper triangular with distinct eigenvalues $\lambda_1, \cdots, \lambda_L$, then 
\begin{align} 
\textstyle e^{[t]} =\sum^{L}_{k=1}  \sum^L_{i=1} \sum^L_{j=i} e_i    p_{i,j,k}  p^{[0]}_j c_k \lambda_k^{t-1}.
\end{align} 
\end{theorem} 

The proof of this theorem is almost the same as that of ~\cite[Theorem 1]{he2016analytic} just with   minor   notation change. 

Theorem~\ref{theorem5} is a special case of Theorem~\ref{theorem2} because distinct eigenvalues $\lambda_1, \cdots, \lambda_L$  means matrix $\mathbf{R}$ is diagonalizable.

\begin{example}[EA-OBSE on OneMax]
\label{exaOneMax1}
 Consider the problem of maximizing the OneMax function,  
\begin{align*} 
\max f  (x)= |x|,  &&x \in \{0,1\}^n.
\end{align*}
EA-OBSE, a (1+1) EA with onebit mutation and strictly elitist selection (Algorithm~\ref{EA-OBSE}),   is used for solving the above maximization problem.

\begin{algorithm}[ht]
\caption{EA-OBSE}
\label{EA-OBSE}
\begin{algorithmic} 
\STATE \textbf{Onebit Mutation:} choose one bit of $x$ at random and flip it.
\STATE \textbf{Elitist Selection}: if $f(y)>f(x)$, then $y$ replace $x$.
\end{algorithmic}
\end{algorithm} 

Let index $i$ denote the state of  $x$ such that $|x|=n-i$ where $i=0, \cdots, n$. The error $e(i)=n-i$. Then transition probabilities satisfy
\begin{align}
\begin{array}{ccc}
    p_{0,0}=1, 
    &p_{i,i+1}= \frac{i+1}{n},
    &p_{i,i}=1-\frac{i}{n} .
\end{array}
\end{align}

Transition matrix $\mathbf{P}$ is upper-triangular. Its power factors, $[p_{i,j,k}]$ (where $i,j,k=1, \cdots, L$), are calculated as follows:
\begin{align} 
p_{i,j,k}=\left\{\begin{array}{lll}
1-\frac{j}{n},  &\mbox{if }i=j=k, 
\\
 0,                &\mbox{if }k<i \mbox{ or } k >j, 
\\
\frac{j}{j-k}  p_{i,j-1,k}, &\mbox{if }i \le k< j, 
\\
 \frac{j}{n}- p_{i,j,j-1}, &\mbox{if }i=j-1 \textrm{ and } j=k,\\
 -\sum^{j-1}_{l=i} p_{i,j,l}, &\mbox{if }i<j-1 \textrm{ and } j=k. 
\end{array}
\right.
\end{align}

Given an initial distribution $\mathbf{p}^{[0]}$, according to Theorem~\ref{theorem5}, the approximation error  \begin{align}
\label{equExample2Error} 
 e^{[t]} =\sum^{L}_{k=1}  \sum^L_{i=1} \sum^L_{j=i} (n-i)    p_{i,j,k}  p^{[0]}_j \left(
 1-\frac{k}{n}\right)^{t-1}.
\end{align} 
(\ref{equExample2Error}) is a closed-form expression of $ e^{[t]}$, which  contains constants, variables,  elementary arithmetic operation ($+, -, \times, \div $) and finite sums. It can be simplified to $e^{[t]}=\left(1-\frac{1}{n}\right)^t e^{[0]}$.  The expression is also given by a much simpler method in Example~\ref{exaOneMax2}.
\end{example}

\begin{example}[EA-OBSE on Mono]
\label{exaMono}
 Consider EA-OBSE for maximizing a monotonically increasing function,  
\begin{align}
&\max f (x),  &x \in \{0,1\}^n,
\end{align}
where $f(x)$ satisfies $f(|x|) <f(|y|)$ if $ |x| < |y| $; $ f(|x|) =f(|y|)$ if $ |x| = |y|$.

Let index $i$ denote the state of  $x$ such that $|x|=n-i$ where $i=0, \cdots, n$. The error $e(i)=f(n)-f(n-i)$. The transition matrix $\mathbf{P}$ is  the same as the above example. Similarly,  the approximation error  
\begin{align*} 
 e^{[t]} =\sum^{n}_{k=1}  \sum^n_{i=1} \sum^n_{j=i} [f(n)-f(n-i)]     p_{i,j,k}  p^{[0]}_j \left(
 1-\frac{k}{n}\right)^{t-1}.
\end{align*}  
Table~\ref{tab1} shows the exact expression of $f^{[t]}$ and $e^{[t]}$ on   $f(x)=|x|$, $|x|^2$ and $\log(|x|+1)$ when $n=4$ and $X^{[0]}=(0000)$. Note that coefficients vary on these  functions. Some coefficients are positive and some are negative. 

\begin{table*}[htb]
\caption{Exact expression  of $f^{[t]}$ and $e^{[t]}$ where $n=4$ and $X^{[0]}= (0000)$ in Example~\ref{exaMono}~\cite{he2016average}.}
\label{tab1}
\centering
\begin{tabular}{c|c }
\toprule
 function $f$  &   $f^{[t]}$
\\ 
\midrule
  $|x|$ &  $4\times(1-0.75\times0.75^{t-1})$ 
  \\ 
  \midrule
 $|x|^2$ &  $16\times(1-1.313\times0.75^{t-1}+0.375 \times0.5^{t-1})$   
\\ \midrule
 $\ln (|x|+1)$ &  $\ln 5\times(1-0.416\times0.75^{t-1}-0.120 \times0.5^{t-1}-0.033\times0.25^{t-1})$   
\\\midrule
\midrule   &   $e^{[t]}$
\\ 
\midrule
  $|x|$ &  $0.75\times0.75^{t-1}$ 
  \\ \midrule 
 $|x|^2$ &  $1.313\times0.75^{t-1}-0.375 \times0.5^{t-1}$   
\\ \midrule
 $\ln (|x|+1)$ &  $0.416\times0.75^{t-1}+0.120 \times0.5^{t-1}+0.033\times0.25^{t-1}$   
\\  
\bottomrule
\end{tabular}
\end{table*}
\end{example}  

If matrix $\mathbf{R}$ is not upper triangular,   Schur's triangularisation states that $\mathbf{R}$ is unitarily similar to an upper triangular matrix. 
\begin{lemma}
\label{lemma6} Let $ \mathbf{R}  = \mathbf{U}^*\mathbf{T} \mathbf{U}$  be Schur's triangularisation of matrix $\mathbf{R}$, where $\mathbf{U}$   is a unitary matrix 
and $\mathbf{T}=[t_{i,j}]$   an upper triangular matrix. Then  
\begin{align} 
\label{equShur}
 e^{[t]} 
 =   \mathbf{e}^T \mathbf{U}^* \mathbf{T}^t \mathbf{U} \mathbf{p}^{[0]}    .
\end{align}  
\end{lemma}

\begin{IEEEproof}
From  $\mathbf{R}  = \mathbf{U}^*\mathbf{T} \mathbf{U}$ and $\mathbf{U}^* \mathbf{U}=\mathbf{I}$ where $\mathbf{I}$ is a unit matrix, we get $\mathbf{R}^t =\mathbf{U}^* \mathbf{T}^t \mathbf{U}$. Inserting it into  (\ref{equApproximationError}), we get the desired conclusion.
\end{IEEEproof}

We need to express the matrix power $\mathbf{T}^t$ in (\ref{equShur}). For any upper triangular matrix $\mathbf{T}$, its power $\mathbf{T}^t$  can be expressed by the entries of $\mathbf{T}$~\cite{huang1978efficient,shur2011simple}.  The following theorem is based on~\cite{dowler2013bounding}.

\begin{theorem} 
\label{theorem6}
Let $ \mathbf{R}  = \mathbf{U}^*\mathbf{T} \mathbf{U}$  be Schur's triangularisation of matrix $\mathbf{R}$, then  
\begin{align} 
\label{equExplicitError1}
 e^{[t]} 
 =  \textstyle \sum^L_{i=1} a_i t^{[t]}_{ij} b_j .
\end{align}  
where vectors $\mathbf{a}^T=[a_i]$ is $\mathbf{e}^T \mathbf{U}^*$ and $\mathbf{b}=[b_i]$ is $\mathbf{U} \mathbf{p}^{(0)}$. $t^{[t]}_{ij}$ is the $ij$th entry of matrix $\mathbf{T}^t$   given by 
\begin{equation}
\label{equRij1}
\small
t^{[t]}_{i,j} =
\left\{
\begin{array}{lll}
\lambda_i^t, \quad \mbox{if } i=j, \\
 \scriptstyle \displaystyle  \sum^{j-i}_{m=1} \sum_{\alpha_i \in A_{[l]}}  \left(\prod^m_{l=1}  t_{\alpha_l \alpha_{l+1}} \sum_{\beta_k \in B_{[m]}^{[t-m]}} \prod^{m+1}_{k=1}  \lambda_{\alpha_k}^{\beta_k}\right), \\
  \qquad \quad  \mbox{if }i <j \\
  0, \qquad \mbox{if } i >j.
\end{array}
\right.
\end{equation}  
where  $\lambda_i =t_{i,i}$ is an eigenvalue of matrix $\mathbf{R}$. The index set $A_{[l]}=\{\alpha_1, \cdots, \alpha_{l+1}\}$ where indexes $\alpha_i$ are positive integers and satisfy $i=\alpha_1 < \alpha_2 < \cdots <\alpha_{l+1}=j$. The index set $B_{[m]}^{[n]}=\{\beta_1, \cdots, \beta_{m+1} \}$ where   indexes $\beta_i$ are  non-negative integers and their sum satisfies $\sum_k \beta_k =n$.
\end{theorem} 

\begin{IEEEproof}
Since matrix  $\mathbf{T}$ is upper triangular, applying~\cite[Theorem 2.4]{dowler2013bounding} to $\mathbf{T}$, we get the expression (\ref{equRij1}) of $t^{[t]}_{i,j}$.   Inserting (\ref{equRij1}) to  (\ref{equShur}), we have the desired result.
\end{IEEEproof}

The above theorem  gives another exact but complicated expression of $e^{[t]}$. Because of Schur's triangularisation, it is hard to generate an exact expression of $e^{[t]}$ in practice.  
  
Summarizing this section, we have demonstrated the exact expression  of $e^{[t]}$ through two methods, albeit the difficulty in obtaining Jordan form and Schur's triangularisation.

\section{Upper Bounds on Approximation Error}
\label{secUpper}
For many EAs, it is complex to obtain an exact expression of $e^{[t]}$. Therefore, a more reasonable goal is to seek an upper bound on $e^{[t]}$. A lower bound on  $e^{[t]}$ is less interesting  because a trivial lower bound always exists: $e^{[t]} \ge 0$. 

\subsection{Convergence Rate Method}
\label{secUpper0}
Unlike an exact expression of $e^{[t]}$, it is rather simple to obtain an upper bound on $e^{[t]}$. A trivial upper bound is 
\begin{align}
    e^{[t]} \le \max \{ e(i);  i =0,\cdots, L\}.
\end{align}
Of course, this upper bound is  loose and unsatisfied.
A better upper bound can be derived from the convergence rate of EAs. 

\begin{definition}
Given an error  sequence $\{e^{[0]}, e^{[1]}, \cdots\}$, its normalized  convergence rate is
$1-e^{[t]}/e^{[t-1]}$ if $e^{[t-1]}\neq 0.$   
\end{definition}

The above rate takes value from $(-\infty, 1]$ and it can be regarded as the convergence speed. The larger value, the faster convergence. Based on this rate, we get an upper bound on $e^{[t]}$. The theorem below is similar to~\cite[Theorem 2]{rudolph1997convergence} but its calculation is more accurate.

\begin{theorem}\label{theorem7}
Given an error sequence $\{e^{[0]}, e^{[1]}, \cdots\}$, define drift $\Delta e(i) = \sum^L_{k=0} p_{k,i}  [f(i)-f(k)]$ where $i=0, \cdots, L$.  If $\Delta e(i)/e(i)>0$ for any $i \neq 0$, then  
\begin{align}
e^{[t]} \le e^{[0]} \left[1-\min_{i=1, \cdots, L} \textstyle\frac{\Delta e(i)}{ e(i)}\right]^t.
\end{align} 
\end{theorem}

\begin{IEEEproof}
We assume that $X^{[t]}=i$ where $i \neq 0$. From 
$$e^{[t]}=e^{[t-1]}- \Delta e(i) \le e^{[t-1]}-\Delta e(i),$$ 
we get
\begin{align}
    \frac{e^{[t]}}{e^{[t-1]}}\le 1 - \frac{\Delta e(i)}{ e(i)} \le 1 -\min_{i=1, \cdots, L} \frac{\Delta e(i)}{ e(i)}.
\end{align}
 
We have
$$\frac{e^{[t]}}{e^{[t-1]}}\le 1 -\min_{i=1, \cdots, L} \frac{\Delta e(i)}{ e(i)},$$ then get the required result.  
\end{IEEEproof}

Let $$
\lambda':=1-\min_{i=1, \cdots, L} \frac{\Delta e(i)}{ e(i)}.$$ 
We show that $\lambda'\ge \lambda_{\max}$ where $\lambda_{\max}$ is the spectral radius of matrix $\mathbf{R}$. For the sake of analysis, we assume $\mathbf{R}>0$\footnote{The analysis of $\mathbf{R}\ge 0$ is similar. The proof needs an extended Collatz-Wielandt formula~\cite[p. 670]{meyer2000matrix}. We omit it in this paper.}. From
$$\textstyle
1-\frac{\Delta e(i)}{ e(i)}=\frac{[\mathbf{e}^T \mathbf{R}]_i}{[\mathbf{e}]_i},$$
according to   the Collatz-Wielandt formula~\cite[p. 669]{meyer2000matrix}, we get 
$$\lambda_{\max}=
\min_{ \mathbf{e}: \mathbf{e}> 0} \max_{i: 1\le i \le L} \frac{[\mathbf{e}^T \mathbf{R}]_i}{e_i},$$
Then
$\lambda'\ge \lambda_{\max}$. Thus $\lambda'$ can be written as $\lambda'=\lambda_{\max}+\epsilon$ for some non-negative $\epsilon$. The above theorem implies 
\begin{align}\textstyle
e^{[t]} \le   c (\lambda_{\max}+\epsilon)^t.
\end{align}

It is worth mentioning that multiplicative drift analysis~\cite{doerr2012multiplicative} also applies the convergence rate $e^{[t]}/e^{[t-1]}  \le \lambda$  to estimating the  hitting time, $\min \{t; e^{[t]}=0\}$. However, multiplicative drift analysis and approximation error analysis discuss two different topics. The former aims at an upper-bound on hitting time while the latter at an  upper bound on approximation error. 

The convergence rate provides a simple method of estimating $e^{[t]}$ and $f^{[t]}$. Its applicability is shown through several examples.

\begin{example}[EA-OBSE on OneMax]
\label{exaOneMax2}
Consider EA-OBSE on the OneMax function. Let $i$ denote the state of $x$ such that $|x|=n-i$. Then $e(i)=i$. We assume that $e^{[t-1]}=i$ where $i>0$, that is,  $X^{[t-1]}$ includes $i$ zero-valued bits. The probability of $X^{[t]}$ reducing $1$ zero-valued bit is $i/n$. 
 \begin{align}
     &\textstyle \Delta e(i)=\frac{i}{n}.\\
     &\textstyle  \frac{\Delta e (i)}{e(i)}= \frac{1}{n}.
 \end{align}
  Then we get
 \begin{align}
     &\textstyle  e^{[t]} =\left(1-\frac{1}{n}\right)^t e^{[0]}.
     \label{equOneMaxE}\\
     &\textstyle f^{[t]} =n-e^{[0]}\left(1-\frac{1}{n}\right)^t.
     \label{equOneMaxF}
 \end{align}
(\ref{equOneMaxF}) is  the same as the result given by~\cite[Theorem 4]{jansen2014performance}.
\end{example}
   
In (\ref{equOneMaxE}), we observe that factor $1-1/n$ equals to the spectral radius $\lambda_{\max}$ of matrix $\mathbf{R}$. In other words,   $e^{[t]} = (\lambda_{\max})^t e^{[0]}.$ This observation is not strange. More generally,  according to~\cite[Theorem 1]{he2016average}, we have a similar result for any EA.

\begin{theorem} 
\label{theorem8}
Let $\lambda_{\max}$ denote the  spectral radius of  matrix $\mathbf{R}$. Choose the initial distribution  $\mathbf{p}^{[0]}= \mathbf{v}$ where $\mathbf{v}$ is an eigenvector corresponding to $\lambda_{\max}$ such that $0<\parallel \mathbf{v} \parallel_1\le 1$, then for any  $t$,
 \begin{align}
e^{[t]} = (\lambda_{\max})^t e^{[0]}.
\end{align} 
\end{theorem}

Notice that in the above example,   ${\Delta e (i)}/{e(i)}=  {1}/{n}$ is independent on $i$. It leads to an exact expression of $e^{[t]}$. But this luck   doesn't always happens. For example, if we change the fitness function a little, say  $f(x) =|x|^2$. In this case $e(i)=n^2-(n-i)^2$. For $i=1, \cdots, n$,
 \begin{align*}\begin{array}{lll}
     &\Delta e(i)=\frac{i}{n} [(n-i+1)^2-(n-i)^2].\\
     &\frac{\Delta e (i)}{e(i)}= \frac{i}{n} \frac{2n-2i+1}{2in-i^2}=\frac{2n-2i+1}{2n^2-ni}.
 \end{array}
 \end{align*}
 Becuase  ${\Delta e (i)}/{e(i)}$ depends on $i$, we cannot get an exact expression of $e^{[t]}$ from Theorem~\ref{theorem7}. 
 
\begin{example}[EA-BWSE on OneMax]
Consider EA-BWSE on the OneMax function. Transition probabilities satisfy
\begin{align}
    &p_{0,0}=1, \\
    &\textstyle \sum_{i: i<j}p_{i,j}\ge \frac{1}{n}\left(1-\frac{1}{n}\right)^{n-1}.
\end{align}
We assume that $X^{[t-1]}=i$. Then 
\begin{align}
     \Delta e(i) &\textstyle \ge \frac{i}{n}\left(1-\frac{1}{n}\right)^{n-1}. \\
    \textstyle 
     \frac{\Delta e (i)}{e(i)} &\textstyle \ge\frac{1}{n}\left(1-\frac{1}{n}\right)^{n-1}  \ge \frac{1}{ne}.
 \end{align}
 Then we get
 \begin{align}
     \textstyle 
     e^{[t]}\le \left(1-\frac{1}{en} \right)^t e^{[0]}.  
 \end{align} 
\end{example}

\begin{example}[EA-BWSE on LeadingOnes]
Consider EA-BWSE for maximizing the LeadingOnes function. 
\begin{align}\textstyle
    f(x) =\sum^n_{i=1} \prod^i_{j=1} x_i.
\end{align}
Let $i$ denote the state of $x$ with $f(x)=n-i$. Then $e(i)=i$.
Transition probabilities satisfy
\begin{align}
    &p_{0,0}=1, \\
    & \textstyle \sum_{i: i<j} p_{i,j}\ge \frac{1}{n}\left(1-\frac{1}{n}\right)^{n-1}.
\end{align}
We assume that $X^{[t-1]}=i$ where $i\ge 1$. Then 
\begin{align}
      \Delta e(i) & \textstyle \ge \frac{1}{n}\left(1-\frac{1}{n}\right)^{n-1}.\\ 
    \textstyle \frac{\Delta e(i)}{e(i)} & \textstyle \ge  \frac{1}{ni}\left(1-\frac{1}{n}\right)^{n-1}\ge  
     \frac{1}{en^2}.
\end{align}
So,
\begin{align}
    &\textstyle e^{[t]} \le \left(1-\frac{1}{en^2}\right)^t e^{[0]} . \\
     &\textstyle f^{[t]} \ge n-e^{[0]} \left(1-\frac{1}{en^2}\right)^t.\label{equLeading1}
\end{align}

Fixed budget analysis has been also applied to EA-BWSE on LeadingOnes~\cite{jansen2014performance}. According to Theorem~\cite[Theorem 13]{jansen2014performance}, if $X^{[0]}$ is chosen uniformly at random, $t =(1-\beta)n^2 /\alpha(n)$ for any $\beta$ with $(1/2)+\beta'<\beta<1$ where $\beta'$ is a positive constant and $\alpha (n)=\omega(1)$, $\alpha (n)\ge 1$, then
\begin{align}
 \label{equLeading2}
\textstyle f^{[t]} = 1+\frac{2t}{n} -o(\frac{t}{n}).
\end{align}

When $X^{[0]}$ is chosen uniformly at random, (\ref{equLeading1}) and (\ref{equLeading2}) are in the same order $1+\Omega(\frac{t}{n})-o(\frac{t}{n})$ for  small $t$. 
However, there exists an essential difference between (\ref{equLeading1}) and (\ref{equLeading2}). (\ref{equLeading1}) is  valid for all $t$ while  (\ref{equLeading2}) is valid for  small $t$ but invalid for  large $t$. Furthermore, the convergence rate method is much simpler than fixed budget analysis in this example~\cite{jansen2014performance}.
\end{example}

\subsection{Auxiliary Matrix Iteration Method}
\label{secUpper1}
An alternative method for upper-bounding the approximation error is to construct an auxiliary matrix iteration converging slower than the original one: $\mathbf{e}^T \mathbf{R}^t \mathbf{q}^{[0]}$. Similar idea has been used in estimating  upper bounds on the computational time of EAs~\cite{he2002individual}.

The meaning of ``slow'' is formalized as follows. Let $\mathbf{A}=[a_{i,j}]$ and $\mathbf{A}'=[a'_{i,j}]$ (where $i,j=0, \cdots, L$) be two non-negative matrices such that $a_{0,0}=1, a_{i,0}=0$ and $a'_{0,0}=1, a'_{i,0}=0$. 
Denote  sub-matrices within states $\{1, \cdots, L\}$ by
$
   \mathbf{B}=[a_{i,j}],  $ $\mathbf{B}'=[b'_{i,j}] $ (where $i,j=1, \cdots, L).
$

\begin{definition} 
The matrix iteration associated with $\mathbf{A}'$   is called \emph{slower} than that with  $\mathbf{A}$,  denoted by $\mathbf{A}' \succeq \mathbf{A}$ or $\mathbf{B}' \succeq \mathbf{B}$, if for any $t \ge 0$, 
\begin{align}
\label{equDominate}
\hat{\mathbf{T}} (\mathbf{B}')^t  \ge \hat{\mathbf{T}} (\mathbf{B})^t, &
\end{align} 
where $L\times L$ matrix $\hat{\mathbf{T}}$ is  upper triangular,  defined by
\begin{align} 
\label{matrixT}
\hat{\mathbf{T}} =  \begin{pmatrix}  
 1    & 1& 1& \cdots &   1    \\ 
    & 1 & 1& \cdots &1 \\
  &    & 1 &  \cdots & 1  \\
  &  &    & \ddots &   \vdots  \\
  &   &   &  &    1     \\
 \end{pmatrix}.
\end{align} 
\end{definition} 

Matrix $\hat{\mathbf{T}}$ comes from the observation that any non-negative vector $\mathbf{e}$ such that $e_1\le  \cdots \le e_L$ equals to 
\begin{equation}
\label{equHatT}
    \mathbf{e}^T =(e_1,   e_{2}-e_{1}, \cdots, e_L-e_{L-1} ) \hat{\mathbf{T}}.
\end{equation}
Let $\hat{\mathbf{e}}^T =(e_1,   e_{2}-e_{1}, \cdots, e_L-e_{L-1} )$ and 
\begin{align}
\label{equError3}
e^{[t]}= \hat{ \mathbf{e}}^T \hat{\mathbf{T}} \mathbf{B}^t \mathbf{p}^{[0]}, &&(e')^{[t]}= \hat{ \mathbf{e}}^T \hat{\mathbf{T}} (\mathbf{B}')^t \mathbf{p}^{[0]}.
\end{align}
From $\mathbf{B}' \succeq \mathbf{B}$, we get that  $(e')^{[t]} \ge e^{[t]} $ for any   $\hat{\mathbf{e}}^T\ge 0$ and $\mathbf{q}^{[0]}\ge 0$.

For any EA, we  draw a general upper bound on $e^{[t]}$ using the auxiliary matrix iteration method.  
\begin{theorem}
\label{theorem3}
Given the matrix $\mathbf{R}$ associated with an EA, let $\lambda_1, \cdots, \lambda_L$  be its eigenvalues. Then there exist some coefficients $c_i$ and small non-negative numbers $\epsilon_i$ such that $e_t$ is upper-bounded by 
\begin{align}
\label{equGeneralUpperBound}
\textstyle    e_t \le \sum^L_{i=1} c_i (\lambda_i+\epsilon_i)^t.
\end{align} 
\end{theorem}

\begin{IEEEproof}
We  choose appropriate $L$ non-negative numbers $\epsilon_1, \cdots, \epsilon_L$ so that  $\lambda_i+\epsilon_i$ are mutually different.  In fact $\epsilon_i$ can be chosen close to $0$. Let $\mathbf{D}$ be the diagonal matrix where $\epsilon_i$ is its $i$th diagonal entry. Denote $\mathbf{R}'=\mathbf{R} + \mathbf{D}$ and $(e')^{[t]} =\mathbf{e}^T (\mathbf{R}')^{t} \mathbf{p}^{[0]}$. Since $\mathbf{R}' \ge  \mathbf{R}$ and   $\hat{\mathbf{T}} (\mathbf{R}')^t \ge  \hat{\mathbf{T}} (\mathbf{R})^t$, we have for any $t$, $e^{[t]} \le (e')^{[t]}.$

Since $\mathbf{R}'$ has $L$ different eigenvalues, it is diagonalizable. According to Theorem~\ref{theorem2},  there exist coefficients $c_i$ such that 
\begin{align}
  \textstyle (e')^{[t]} = \sum^L_{i=1} c_i (\lambda_i+\epsilon_i)^t.
\end{align}
From $e^{[t]} \le (e')^{[t]}$, we get  the desired conclusion. 
\end{IEEEproof}

The above theorem  also gives a guideline for seeking a lower  bound on $f^{[t]}$, that is  $f^{[t]}\ge f_{\opt} -\sum^L_{i=1} c_i (\lambda_i+\epsilon_i)^t$. 

Obviously it is inconvenient to verify (\ref{equDominate}) for each $t$. Hence it is necessary to seek  conditions of determining whether (\ref{equDominate}) is true.  
Theorem~\ref{theorem9} below provides such a sufficient condition. 
 
\begin{theorem}
\label{theorem9}
If  non-negative matrices $\mathbf{B}'$   and  $\mathbf{B}$   satisfy two conditions:
\begin{align}  
&\hat{\mathbf{T}} \mathbf{B}'\ge \hat{\mathbf{T}}  \mathbf{B},\label{Condition1}\\
&\hat{\mathbf{T}} \mathbf{B}' \hat{\mathbf{T}}^{-1}\ge 0, \label{Condition2}  
\end{align} then $\mathbf{B}' \succeq \mathbf{B}$.
\end{theorem}

\begin{IEEEproof}  
When $t=0$,   (\ref{equDominate}) is trivial. 
When $t=1$,   (\ref{equDominate}) is derived from  (\ref{Condition1}).
Now we assume   (\ref{equDominate})  is true  for some $t \ge 0$.
Let's prove (\ref{equDominate}) is true for $t+1$. 

Since
\begin{align*}
(\hat{\mathbf{T}} \mathbf{B}' \hat{\mathbf{T}}^{-1})  \hat{\mathbf{T}}  (\mathbf{B}')^{t} &\ge( \hat{\mathbf{T}} \mathbf{B}' \hat{\mathbf{T}}^{-1}) \hat{\mathbf{T}}  (\mathbf{B})^{t}.  && (\mbox{use (\ref{Condition2}))}\\
\hat{\mathbf{T}}   (\mathbf{B}')^{t+1} &\ge \hat{\mathbf{T}} \mathbf{B}'   (\mathbf{B})^{t}\\
& \ge \hat{\mathbf{T}} \mathbf{B}   (\mathbf{B})^{t} =\hat{\mathbf{T}}   (\mathbf{B})^{t+1}. && (\mbox{use (\ref{Condition1}))}
\end{align*}
This proves that (\ref{equDominate}) is true for $t+1$. 
By induction,  (\ref{equDominate}) is true for any $t \ge 0$.
\end{IEEEproof}

The above theorem provides  sufficient conditions that an auxiliary chain is slower than the original one.  
 
We apply Theorem~\ref{theorem9} to elitist EAs. The  transition matrix of elitist EAs is either upper-triangular or block upper-triangular matrices. For an upper-triangular matrix $\mathbf{R}$, the theorem below show a  method of constructing matrix $\mathbf{R}'$ such that $\mathbf{R}' \succeq \mathbf{R}$.

\begin{theorem}
\label{theorem10} 
Provided that transition matrix $\mathbf{P}$ is upper triangular, construct another transition matrix $\mathbf{P}'$ which is upper triangular and  satisfies 
\begin{align}
\label{conC1}
&p'_{j,j}  \ge  p_{j,j}, &&\textrm{for any } j,
\\
\label{conC2}
&\textstyle \sum^{i-1}_{l=0} (p_{l,j}-p'_{l,j}) \ge 0 , &&\textrm{for  any }i<j, 
\\
\label{conC3}
&\textstyle  \sum^{i}_{l=0} ( p'_{l,j-1}- p'_{l,j})\ge 0 , &&\textrm{for any }i<j-1.
\end{align}  
then   $\mathbf{R}' \succeq \mathbf{R}$.
\end{theorem} 

\begin{IEEEproof}
For   matrices $\mathbf{R}'$ and $\mathbf{R}$,  we prove   they satisfy the conditions of Theorem~\ref{theorem9}. 
  
Let $\hat{\mathbf{T}}$  be the special upper-triangular matrix given by~(\ref{matrixT}). First we verify that $\hat{\mathbf{T}}  \mathbf{R}' \ge \hat{\mathbf{T}} \mathbf{R}$. Matrix $\hat{\mathbf{T}} \mathbf{R}'$ equals to
 \begin{align*} 
 \begin{pmatrix}  
  p'_{1,1}    & \sum^{2}_{i=1} p'_{i,2}&   \cdots &  \sum^{L-1}_{i=1} p'_{i,L-1}  &\sum^L_{i=1} p'_{1,L}  \\ 
   & p'_{2,2} &    \cdots &   \sum^{L-1}_{i=2} p'_{i,L-1}  &\sum^L_{i=2} p'_{i,L} \\ 
   &   & \ddots & \vdots   & \vdots  \\
    &   &      &     p'_{L-1,L-1}  &\sum^L_{i=L-1} p'_{i,L} \\ 
 &   &    &  & p'_{L,L}     \\
 \end{pmatrix}.
\end{align*}
Its $ij$-th entry (where $i< j$)  is 
\begin{align}  
\label{equEntry1}
\textstyle \sum^{j}_{l=i} p'_{l,j}  =  1-\sum^{i-1}_{l=0} p'_{l,j}.
\end{align}

Similarly, the $ij$th entry of $\hat{\mathbf{T}} \mathbf{R}$ (where $i < j$) is
\begin{align}
\label{equEntry2}
\textstyle \sum^{j}_{l=i} p_{l,j} 
 = 1-\sum^{i-1}_{l=0} p_{l,j}.
\end{align}

For the $jj$th entry,   (\ref{conC1}) states 
$
p'_{j,j}  \ge  p_{j,j}.
$
 
For the $ij$th entry with $i<j$,  from (\ref{equEntry1}) and (\ref{equEntry2}), we have
\begin{align*}
&\textstyle \sum^{j}_{l=i} p'_{l,j} -
\sum^{j}_{l=i} p_{l,j}  \\
=& \textstyle  (1-\sum^{i-1}_{l=0} p'_{l,j})
- (1-\sum^{i-1}_{l=0} p_{l,j})\\
=& \textstyle \sum^{i-1}_{l=0} (p_{l,j}-p'_{l,j}) \ge 0 \qquad (\mbox{use (\ref{conC2})}).
\end{align*} 

Then  we come to $\hat{\mathbf{T}} \mathbf{R}' \ge \hat{\mathbf{T}} \mathbf{R}.$
 
Secondly, we verify that  $\hat{\mathbf{T}} \mathbf{R}' \hat{\mathbf{T}}^{-1}\ge 0$.  From
\begin{align} 
\hat{\mathbf{T}}^{-1} =  \begin{pmatrix}  
 1    & -1& 0& \cdots & 0    & 0\\ 
   & 1 & -1& \cdots &0 &0\\
  &   & 1 &  \cdots & 0 &0\\
   &   &    & \ddots & \vdots  &\vdots \\
  &   &   &   & 1   &-1 \\
 &   &  &  &  & 1     \\
 \end{pmatrix},
\end{align} we get that matrix $\hat{\mathbf{T}} \mathbf{R}' \hat{\mathbf{T}}^{-1}$   equals  to
 \begin{align*}  
 \begin{pmatrix}  
  p'_{1,1}    &  \displaystyle-p'_{1,1}+\sum^{2}_{i=1} p'_{i,2}&   \cdots  &\displaystyle -\sum^{L-1}_{i=1} p'_{i,L-1}+ \sum^L_{i=1} p'_{1,L}  \\ 
    & p'_{2,2} &    \cdots   &\displaystyle-\sum^{L-1}_{i=2} p'_{i,L-1} +\sum^L_{i=2} p'_{i,L} \\ 
  &   & \ddots & \vdots     \\ 
  &   &    &  p'_{L,L}     \\
 \end{pmatrix}.
\end{align*}

Because Condition  (\ref{conC1}) states $p'_{j,j}  \ge  p_{j,j}$, we only need to prove that its $ij$-th entry (where $i<j$) is non-negative. This entry equals to
\begin{align*}
&\textstyle\sum^j_{l=i}  p'_{l,j}-\sum^{j-1}_{l=i} p'_{l,j-1}  \\
=&\textstyle (1-\sum^{i-1}_{l=0} p'_{l,j}) -(1-\sum^{i}_{l=0} p'_{l,j-1})\\
=& \textstyle  \sum^{i}_{l=0} ( p'_{l,j-1}- p'_{l,j})\ge 0 \quad (\textrm{use  (\ref{conC3})}). 
\end{align*}
Thus  $\hat{\mathbf{T}} \mathbf{R}' \hat{\mathbf{T}}^{-1}$ is non-negative

According to Theorem~\ref{theorem9}, we know  $\mathbf{R}' \succeq \mathbf{R}$.
\end{IEEEproof} 

A simple way to construct an auxiliary Markov chain is a bidiagonal transition matrix  $\mathbf{P}'$, which is given as follows
\begin{align}
\left\{
\begin{array}{ll}
p'_{0,0} =1,\\
p'_{j-1,j}\le  \sum^{j-1}_{l=0} p_{l,j},\\
p'_{j,j} =1 - p'_{j-1,j},\\
p'_{i,j}=0, &\mbox{others}.
\end{array}
\right.
\end{align} 

For any matrix $\mathbf{R}' \succeq \mathbf{R}$, the convergence rate method is applicable to $\mathbf{R}'$. From Theorem~\ref{theorem7}, we get an upper bound on $e^{[t]}$ as follows.

\begin{theorem}\label{theorem11}
For any matrix $\mathbf{R}' \succeq \mathbf{R}$, define drift $\Delta e'(j) = \sum^L_{i=0} p'_{i,j}  [f(j)-f(i)]$ where $j=1, \cdots, L$.    then  
\begin{align}
e^{[t]} \le e^{[0]} \left[1-\min_{i=1, \cdots, L} \textstyle\frac{\Delta e'(i)}{ e'(i)}\right]^t.
\end{align} 
\end{theorem}

\begin{example}[EA-BWSE on Mono]
\label{exampleMono2}
Consider EA-BWSE for maximizing a monotonically increasing function $f(x)$.  
Let index $i$ stands for the state of $x$ such that $|x|=n-i$  where $i=0, \cdots, n$.  Then error $e(i)=f(n) -f(n-i).$ 

Transition probabilities satisfy
\begin{align}\label{tp}
\begin{array}{lll}
    &p_{0,0}=1, \\
    &    p_{i-1,i}\le \frac{i}{n} \left(1-\frac{1}{n}\right)^{n-i}, &\mbox{if } 1\le i\le n.
\end{array}
\end{align}

Construct an auxiliary upper-triangular transition matrix $\mathbf{P}'$ as follows:
\begin{align}
\begin{array}{lll}
    p'_{0,0}=1, \\
    p'_{i-1,i}=\frac{i}{n} \left(1-\frac{1}{n}\right)^{n-i}, &\mbox{if } 1\le i\le n,\\
    p'_{i,i}= 1-\frac{i}{n} \left(1-\frac{1}{n}\right)^{n-i}.
\end{array}
\end{align} 

We assume that $X^{[t-1]}=i$ where $i\ge 1$. Then 
\begin{align}
      \Delta e'(i) & \textstyle = [f(n-i+1)-f(n-i)] \frac{i}{n}\left(1-\frac{1}{n}\right)^{n-1}.\\ 
      \textstyle \frac{\Delta e'(i)}{e'(i)} &  \ge  \min_{i=1,\cdots, L}\textstyle  \frac{f(n-i+1)-f(n-i)}{f(n)-f(n-i)} \frac{i}{n}\left(1-\frac{1}{n}\right)^{n-1}.
\end{align}
So,
\begin{align*} 
e^{[t]} \le e^{[0]} \left[1-\min_{i=1,\cdots, L} \textstyle  \frac{f(n-i+1)-f(n-i)}{f(n)-f(n-i)} \frac{i}{n}\left(1-\frac{1}{n}\right)^{n-1}\right]^t.
\end{align*}
\end{example}

The upper bound derived by the convergence rate method might be loose.  If transition matrix $\mathbf{P}'$ is upper triangular, the upper bound may be  be improved through the power factor method. A requirement of applying the power factor  method is that diagonal entries of matrix $\mathbf{P}'$ are unique. However this requirement can be easily achieved. If $p'_{i,i}=p'_{j,j}$, we can add  a small positive $\epsilon_i  0$ to $p'_{i,i}$ so that  $p'_{i,i}+\epsilon_i\neq p'_{j,j}$.

\begin{theorem} \label{theorem12} 
Provided that transition matrix $\mathbf{P}$ is upper triangular, construct another upper triangular transition matrix $\mathbf{P}'$ satisfying Conditions~(\ref{conC1}), (\ref{conC2}) and (\ref{conC3}), 
and $\lambda'_j:=p'_{j,j}$ are mutually different. 
The power factors of matrix $\mathbf{R}'$, $[p'_{i,j,k}]$ (where $i,j,k=1, \cdots, L$) are given by (\ref{equPowerFactors}).
Then given an initial distribution $\mathbf{p}^{[0]}$,
\begin{align}
\label{equUpperBound10}
e^{[t]} \le \sum^{n}_{k=1}  \sum^k_{i=1} \sum^n_{j=k} e_i    p'_{i,j,k}  p^{[0]}_j ( \lambda'_k)^{t-1}.
\end{align} 
\end{theorem} 

\begin{IEEEproof}
From Theorem~\ref{theorem10}, we know $e^{[t]} \le (e')^{[t]}$. From Theorem~\ref{theorem5}, we get 
\begin{align}
 e^{[t]} \le (e')^{[t]} =\sum^{n}_{k=1}  \sum^n_{i=1} \sum^n_{j=i} e_i    p'_{i,j,k}  p^{[0]}_j (\lambda'_k)^{t-1}.
\end{align}
 Since $p'_{i,j,k}=0$ if $ k<i$ or $k>j,$
 we know that
 \begin{align}
 e^{[t]} \le \sum^{n}_{k=1}  \sum^k_{i=1} \sum^n_{j=k} e_i    p'_{i,j,k}  p^{[0]}_j (\lambda'_k)^{t-1}.
\end{align}
This  is the wanted conclusion.
\end{IEEEproof}

Recall that $p'_{i,i}  \ge  p_{i,i}$, equivalently   $\lambda'_i \ge \lambda_i$. (\ref{equUpperBound10}) can be rewritten as
\begin{align}\textstyle
e^{[t]} \le \sum^L_{i=1} c_i (\lambda_i+\epsilon_i)^t,
\end{align} 
for some coefficients $c_i$ and non-negative numbers $\epsilon_i$.

\begin{example}[EA-BWSE on Mono]
\label{exampleMono3}
Consider EA-BWSE for maximizing a monotonically increasing function $f(x)$.
Let index $i$ stands for the state of $x$ such that $|x|=n-i$  where $i=0, \cdots, n$. The error $e_{i}=f(n)-f(n-i)$.

Constructor an  auxiliary transition matrix $\mathbf{P}'$  the same  as that in Example~\ref{exampleMono2}. By Theorem \ref{theorem12} we get
\begin{align}\label{Ubound}
 e^{[t]} \le (e')^{[t]} =\sum^{n}_{k=1}  \sum^k_{i=1} \sum^n_{j=k} e_i    p'_{i,j,k}  p^{[0]}_j (\lambda'_k)^{t-1}.
\end{align}
The value of $\lambda'_k$    is
\begin{align*}
\lambda'_k=1-\textstyle \frac{k}{n}  \left(1-\frac{1}{n}\right)^{n-k}, &&k=1, \cdots, n.
\end{align*}
The value of   $p'_{i,j,k}$ is calculated by (\ref{equPowerFactors}) and listed as follows:
\begin{align*}
p'_{1,1,1}&= \textstyle \frac{1}{n}  \left(1-\frac{1}{n}\right)^{n-1},\\
p'_{2,2,2}&= \textstyle  \frac{2}{n}  \left(1-\frac{1}{n}\right)^{n-2},\\
p'_{1,2,1} & =\textstyle \frac{2[n-(1-\frac{1}{n})^{n-1}]}{n+1}. \\
p'_{1,2,2} & =\textstyle -\frac{2[n-2(1-\frac{1}{n})^{n-2}]}{n+1} \cdots
\end{align*}
We omit the full list of  $p'_{i,j,k}$, because it   is lengthy and will distract  analysis.

Given an initial distribution $\mathbf{p}^{[0]}$ with $p^{[0]}_0+\cdots+p^{[0]}_n=1$,  $e^{[t]}$  could be estimated level by level.  For example,
\begin{enumerate}
\item If $p^{[0]}_1=1$,  we have
 \begin{align*}
 e^{[t]}_1 & \le   e_1    p'_{1,1,1}  p^{[0]}_1 (\lambda'_1)^{t-1}\\
 &=\textstyle  \left[f(n)-f(n-1)\right]    \left[1-\frac{1}{n}
 \left(1-\frac{1}{n}\right)^{n-1}\right]^{t}\\
  &\le \textstyle  \left[f(n)-f(n-1)\right]    \left(1-\frac{1}{ne}\right)^{t},
\end{align*}
where $ e^{[t]}_i$   denotes $ e^{[t]}(|X^{[0]}|=n-i)$.

\item If $p^{[0]}_2=1$, we have
\begin{align*}
& e^{[t]}_2 \le \sum^{2}_{k=1}  \sum^k_{i=1}  e_i    p'_{i,2,k}   ( \lambda'_k)^{t-1}\\
=&\scriptstyle e_1p'_{1,2,1}(\lambda'_{1})^{t-1}+\left(e_1p'_{1,2,2}+e_2p'_{2,2,2}\right)(\lambda'_{2})^{t-1}\\
=&\scriptstyle [f(n)-f(n-1)]\left\{\frac{2n}{n+1}\left[\left(1-\frac{1}{n}(1-\frac{1}{n})^{n-1}\right)^{t}-\left(1-\frac{2}{n}(1-\frac{1}{n})^{n-2}\right)^{t}\right]\right\}\\
 &\scriptstyle +[f(n)-f(n-2)]\left(1-\frac{2}{n}(1-\frac{1}{n})^{n-2}\right)^{t}\\
\le &\scriptstyle 2[f(n)-f(n-1)]\left(1-\frac{1}{ne}\right)^t+[f(n)-f(n-2)]\left(1-\frac{2}{ne}\right)^t.
\end{align*} 
\end{enumerate}
\end{example}

Summarizing this section, we propose two  methods for upper-bounding $e^{[t]}$. The convergence rate method is   simple but applicable to all   EAs. The method of  auxiliary matrix iteration + power factors   might provide a better upper bound, but it is more complex and only works on elitist EAs.

\section{Conclusions}
\label{secConclusions} 
This paper establishes a novel theoretical framework of analyzing the approximation error of EAs for discrete optimization.   In this framework, EAs are modelled by homogeneous Markov chains. The framework is divided into two parts. 

The first part is about exact expressions of the approximation error. Two methods, Jordan form and Schur's triangularization, are proposed for studying the exact expression of $e^{[t]}$. 
It is proven that the exact expression of $e^{[t]}$ is  
\begin{align} \textstyle
e^{[t]}  
=   \sum^k_{i=1} \sum^{L_i}_{m=1}   c_{i_m} \binom{t}{L_i-m+1} \lambda^{t-m+1}_{i},
\end{align} 
where  $\lambda_i$ are  eigenvalues of   matrix $\mathbf{R}$, $c_{i_m}$   coefficients, $k$ and $L_i$ are integers. 
If  matrix $\mathbf{R}$ is diagonalizable, then the exact expression of $e^{[t]}$ can be simplified as      
\begin{align}\textstyle
e^{[t]} = \sum^L_{i=1} c_i \lambda_i^t.
\end{align} 

The second part is about upper bounds on the approximation error.  
Two methods, convergence rate and auxiliary matrix iteration,   are introduced to the estimation of  the upper bound on $e^{[t]}$. The convergence rate method is used to derive an upper bound on $e^{[t]}$ in the form
\begin{align}\textstyle
e^{[t]} \le   c (\lambda_{\max}+\epsilon)^t,
\end{align} 
where $\lambda_{\max}$ is the spectral radius of matrix $\mathbf{R}$, $c$ a coefficient and $\epsilon$ a small non-negative number.
If  matrix $\mathbf{R}$ is upper triangular,  the method of auxiliary matrix iteration + power factors gives an upper bound   in  the form
\begin{align}\textstyle
e^{[t]} \le \sum^L_{i=1} c_i (\lambda_i+\epsilon_i)^t,
\end{align}  
where $c_i$ coefficients and $\epsilon_i$ small non-negative numbers. Parameters $\epsilon, \epsilon_i$ could be chosen as small as close to $0$.

The applicability of this framework is demonstrated through several examples. The approximation error analysis of EAs is still at an early stage. Our future work is to apply this framework to more EAs on more problems.


\end{document}